\documentclass[letterpaper, 10 pt, conference]{ieeeconf}  % Comment this line out if you need a4paper

\IEEEoverridecommandlockouts                              % This command is only needed if 
\usepackage{graphics} % for pdf, bitmapped graphics files
\usepackage{epsfig} % for postscript graphics files
\usepackage{mathptmx} % assumes new font selection scheme installed
\usepackage{times} % assumes new font selection scheme installed

\usepackage[T1]{fontenc} % optional
\usepackage{cite}
\usepackage{booktabs}
\usepackage{amsmath,amssymb,amsfonts}
\usepackage{algorithmic}
\usepackage{setspace}
\usepackage{subfigure}
\usepackage{graphicx}
\usepackage{textcomp}
\usepackage{xcolor}
\usepackage{multirow}
\usepackage{threeparttable}
\usepackage{tabularx}
\usepackage{mathtools}
\usepackage[free-standing-units=true]{siunitx}
\newcommand{\rom}[1]{\uppercase\expandafter{\romannumeral #1\relax}}

\usepackage{array}
\newcommand{\PreserveBackslash}[1]{\let\temp=\\#1\let\\=\temp}
\newcolumntype{C}[1]{>{\PreserveBackslash\centering}p{#1}}
\newcolumntype{R}[1]{>{\PreserveBackslash\raggedleft}p{#1}}
\newcolumntype{L}[1]{>{\PreserveBackslash\raggedright}p{#1}}

\newcommand\yuheng[1]{\textcolor{black}{#1}}
\newcommand{\fref}[1]{Fig. \ref{#1}}
\newcommand{\sref}[1]{Section \ref{#1}}
\newcommand{\tref}[1]{Table \ref{#1}}
\newcommand{\argmin}{\operatornamewithlimits{argmin}}

\makeatletter
\let\NAT@parse\undefined
\makeatother

\usepackage[colorlinks]{hyperref}
  \hypersetup{
    citecolor=blue,
    linkcolor=blue,   
    urlcolor=blue}

\author{Yuheng Qiu$^{1}$, Chen Wang$^{1}$, Wenshan Wang$^{1}$, Mina Henein$^{2}$, and Sebastian Scherer$^{1}$% <-this % stops a space
\thanks{*This work was partially sponsored by the Sony award \#A023367 and ARL DCIST CRA award W911NF-17-2-0181.}% <-this % stops a space
\thanks{Source Code: \url{https://github.com/haleqiu/AirDOS}.}% <-this % stops a space
\thanks{$^{1}$Yuheng Qiu, Chen Wang, Wenshan Wang, and Sebastiian Scherer are with the Robotics Institute, Carnegie Mellon University, Pittsburgh, PA 15213, USA {\tt\small \{yuhengq, wenshanw, basti\} @andrew.cmu.edu; chenwang@dr.com}}%
\thanks{$^{2}$Mina Henein is with  the System, Theory and Robotics Lab, Australian National University. {\tt\small mina.henein@anu.edu.au}}%
}

\begin{document}

\title{\LARGE \bf
AirDOS: Dynamic SLAM benefits from Articulated Objects
}

\thispagestyle{empty}
\pagestyle{empty}

\maketitle

\begin{abstract}
Dynamic Object-aware SLAM (DOS) exploits object-level information to enable robust motion estimation in dynamic environments.
% It has attracted increasing attention with the recent success of learning-based models.
Existing methods mainly focus on identifying and excluding dynamic objects from the optimization. In this paper, we show that feature-based visual SLAM systems can also benefit from the presence of dynamic articulated objects by taking advantage of two observations: (1) The 3D structure of each rigid part of articulated object remains consistent over time; (2) The points on the same rigid part follow the same motion. In particular, we present AirDOS, a dynamic object-aware system that introduces rigidity and motion constraints to model articulated objects. By jointly optimizing the camera pose, object motion, and the object 3D structure, we can rectify the camera pose estimation, preventing tracking loss, and generate 4D spatio-temporal maps for both dynamic objects and static scenes. Experiments show that our algorithm improves the robustness of visual SLAM algorithms in challenging crowded urban environments. To the best of our knowledge, AirDOS is the first dynamic object-aware SLAM system demonstrating that camera pose estimation can be improved by incorporating dynamic articulated objects.
\end{abstract}

\section{Introduction}
\begin{figure}[t]
	\centering
    \subfigure[Challenge of Shibuya Tokyo]{
    	\includegraphics[width=0.45\linewidth]{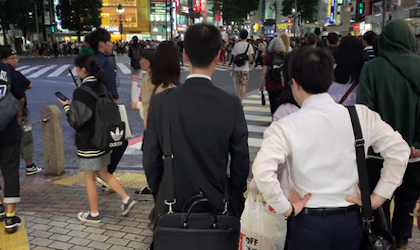}
    	\label{Fig:shibuya}
	}
	\subfigure[TartanAir Shibuya Dataset]{
    	\includegraphics[width=0.45\linewidth]{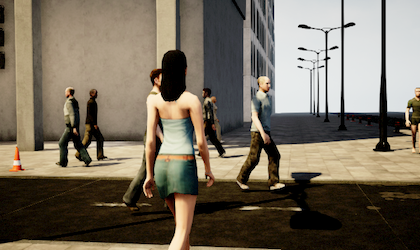}
    	\label{Fig:tartan_shibuya}
	}\\
	\centering
	\subfigure[Example of KITTI tracking dataset training 19]{
    	\includegraphics[width=0.96\linewidth]{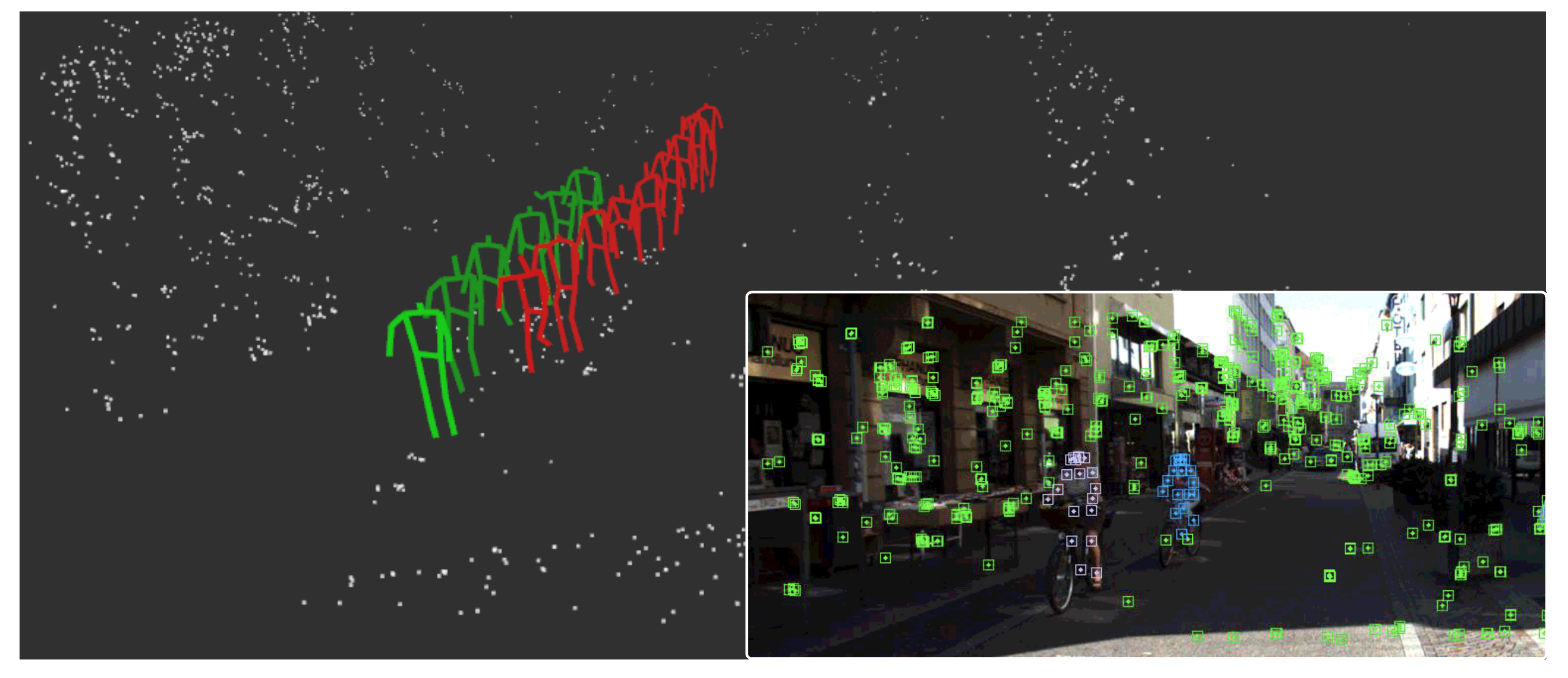}
    	\label{Fig:kitti_result}
	}
	\caption{(a) Example of a highly dynamic environment cluttered with humans which represents a challenge for Visual SLAM.  Existing dynamic SLAM algorithms often fail in this challenging scenario (b) Example of the TartanAir Shibuya Dataset. (c) Example of the estimated full map with dynamic objects and static background.}
	\label{fig:motivation}
	\vspace{-10pt}
\end{figure}

Simultaneous localization and mapping (SLAM) is a fundamental research problem in many robotic applications.
Despite its success in static environments, the performance degradation and lack of robustness in the dynamic world has become a major hurdle for its practical applications
\cite{cadena2016past,wang2017non}.
To address the challenges of dynamic environments, most SLAM algorithms adopt an \textit{elimination strategy} that treats moving objects as outliers and estimates the camera pose only based on the measurements of static landmarks \cite{mur2015orb,engel2017direct}.
This strategy can handle environments with a small number of dynamics, but cannot address challenging cases, where dynamic objects cover a large field of view as in \fref{Fig:shibuya}.

Some efforts have been made to include dynamic objects in the SLAM process. Very few methods try to estimate the pose of simple rigid objects \cite{nicholson2018quadricslam, yang2019cubeslam} or estimate their motion model \cite{henein2018exploiting,judd2018multimotion}.
For example, CubeSLAM \cite{yang2019cubeslam} introduces a simple 3D cuboid to model rigid objects. Dynamic SLAM \cite{henein2020dynamic} estimates 3D motions of dynamic objects.
However, these methods can only cover special rigid objects, e.g., cubes \cite{yang2019cubeslam} and quadrics \cite{nicholson2018quadricslam} and do not show that camera 
 pose estimation can be improved by the introduction of dynamic objects \cite{henein2020dynamic,henein2018exploiting,judd2018multimotion}.
This introduces our main question: 

\textit{Can we make use of moving objects in SLAM to improve camera pose estimation rather than filtering them out? }

% \begin{IEEEkeywords}
% Visual SLAM, Dynamic Mapping, and Rigidity Constraint
% \end{IEEEkeywords}

\begin{figure}[!t]
	\centering
    \includegraphics[width=1\linewidth]{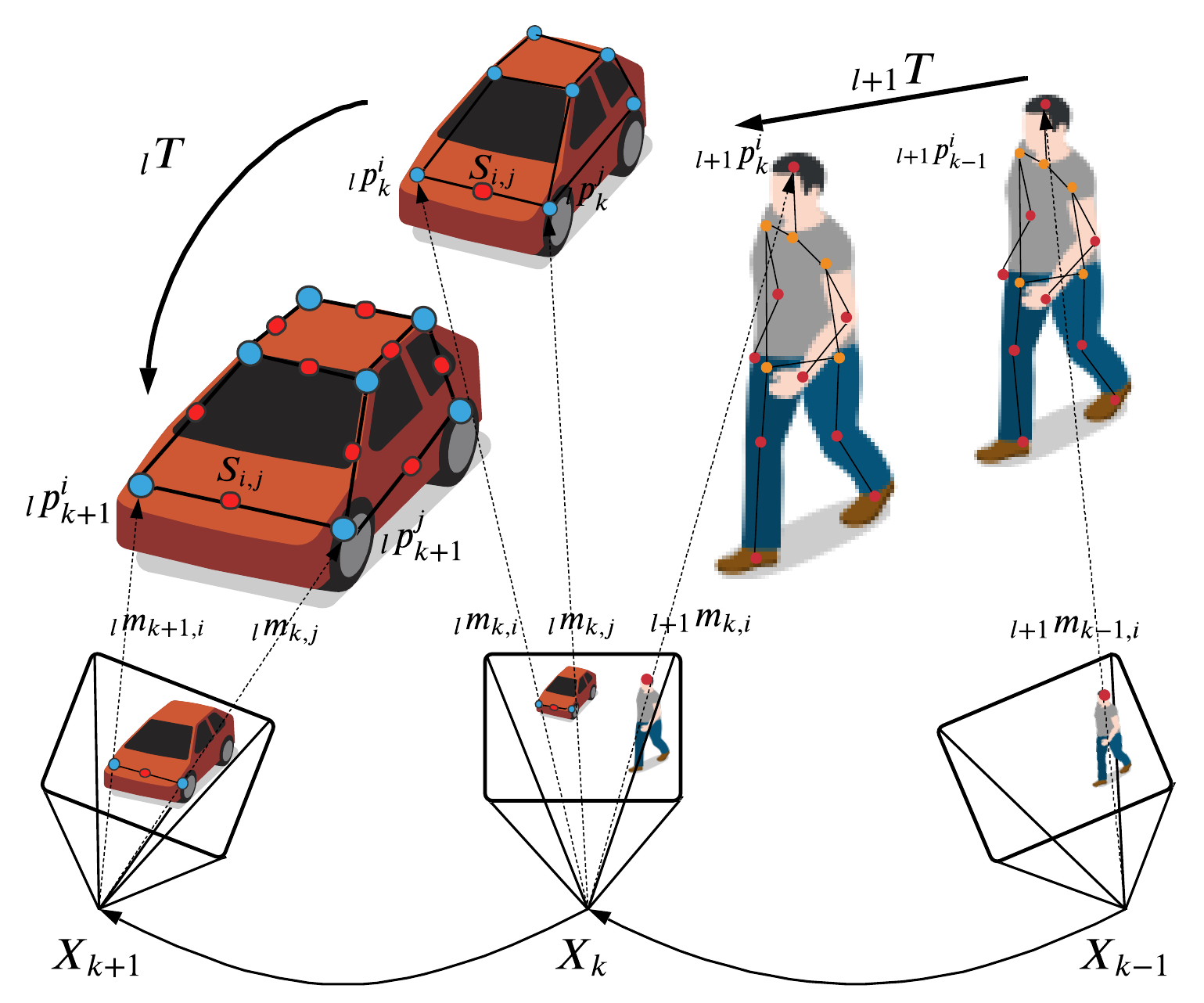}
    \caption{This is an example of the articulated dynamic objects' point-segment mode. In urban environment, we can model rigid objects like vehicle and semi-rigid objects like pedestrian as articulated object. $p_k^i$ and $p_k^j$ are the $i$-th and $j$-th dynamic features on the moving objects at time $k$. $p_{k+1}^i$ and $p_{k+1}^j$ is the dynamic features after the motion $_lT_{k}$ at time $k+1$. In this model, the segment $s^{ij}$ is invariant over time and motion.}
	\label{Fig:point-segment model}
	\vspace{-8pt}
\end{figure}

In this paper, we extend the simple rigid objects to general \emph{articulated objects}, defined as objects composed of one or more rigid parts (links) connected by joints allowing rotational motion \cite{stamou20054}, e.g., vehicles and humans in \fref{Fig:point-segment model}, and utilize the properties of articulated objects to improve the camera pose estimation. Namely, we jointly optimize (1) the 3D structural information and (2) the motion of articulated objects.
% We demonstrate that dynamic objects can not only be beneficial to the mapping, but also to the camera pose estimation.
To this end, we introduce (1) a rigidity constraint, which assumes that the distance between any two points located on the same rigid part remains constant over time, and (2) a motion constraint, which assumes that feature points on the same rigid parts follow the same 3D motion.
This allows us to build a 4D spatio-temporal map including both dynamic and static structures.

In summary, the main contributions of this paper are:
\begin{itemize}
\item A new pipeline, named AirDOS, is introduced for stereo SLAM to jointly optimize the camera poses, trajectories of dynamic objects, and the map of the environment.
% , which improves the robustness and accuracy.
\item We introduce simple yet efficient rigidity and motion constraints for general dynamic articulated objects.
% \item We evaluate AirDOS on KITTI dataset, and introduce a new benchmark TartanAir Shibuya, which demonstrates that, for the first time, dynamic articulated objects benefit the camera pose estimation in SLAM. 
\item \yuheng{We introduce a new benchmark TartanAir Shibuya, on which we demonstrates that, for the first time, dynamic articulated objects can benefit the camera pose estimation in visual SLAM.} 
\end{itemize}

\section{Related Work}
Recent works on dynamic SLAM roughly fall into three categories: elimination strategy,  motion constraint, and rigidity constraint, which will be reviewed, respectively.

\subsection{Elimination Strategy}

Algorithms in this category filter out the dynamic objects and only utilize the static structures of the environment for pose estimation.
Therefore, most of the algorithms in this category apply elimination strategies like RANSAC \cite{fischler1981random} and robust loss functions \cite{kerl2013robust} to eliminate the effects of dynamic objects.
For example, ORB-SLAM \cite{mur2015orb} applies RANSAC to select and remove points that cannot converge to a stable pose estimation.
DynaSLAM \cite{bescos2018dynaslam} detects the moving objects by multi-view geometry and deep learning modules. This allows inpainting the frame background that has been occluded by dynamic objects.
Bârsan et al. \cite{barsan2018robust} use both instance-aware semantic segmentation and sparse scene flow to classify objects as either background, moving, or potentially moving objects.
Dai et al. \cite{dai2020rgb} utilize the distance correlation of map points to segment dynamic objects from static background.
To reduce the computational cost, Ji et al. \cite{ji2021towards} combine semantic segmentation and geometry modules, which clusters the depth image into a few regions and identify dynamic regions via reprojection errors.

\subsection{Motion Constraint}
Most algorithms in this category estimate the motion of dynamic objects but do not show that the motion constraint can contribute to the camera pose estimation, and would thus suffer in highly dynamic environments.
For example, Hahnel et al. \cite{hahnel2003map} track the dynamic objects in the SLAM system.
Wang et al. \cite{wang2007simultaneous} introduce a simultaneous localization, mapping, and moving object tracking (SLAMMOT) algorithm, which tracks moving objects with a learned motion model based on a dynamic Bayesian network.
Reddy, et al. \cite{reddy2015dynamic} use optical flow to segment moving objects, and apply a smooth trajectory constraint to enforce the smoothness of objects' motion.
Judd et al. \cite{judd2018multimotion} propose multi-motion visual odometry (MVO), which simultaneously estimates the camera pose and the object motion.
% However, although the motion and camera pose are estimated simultaneously in MVO, the motion is only determined by the camera pose estimation, in which any failure will lead the motion estimation to fail.
% Henein, and Zhang et al. \cite{minadynamic, zhang2020vdo} generate a map of dynamic and static structure and estimate velocities of rigid moving objects using motion constraints.
The work by Henein, et al. \cite{zhang2020vdo, minadynamic, henein2018exploiting}, of which the most recent is VDO-SLAM \cite{zhang2020vdo}, generate a map of dynamic and static structure and estimate velocities of rigid moving objects using motion constraints.
Rosinol, et al. \cite{rosinol20203dkimera} propose 3D dynamic scene graphs to detect and track dense human mesh in dynamic scenes. This method constraints the humans maximum walking speed for a consistency check.

\subsection{Rigidity Constraint}

Rigidity constraint assumes that pair-wise distances of points on the same rigid body remain the same over time.
It was applied to segment moving objects in dynamic environments dating back to the 1980s.
Zhang et al.\cite{zhang1988analysis} propose to use rigidity constraint to match moving rigid bodies. 
Thompson et al.\cite{thompson1993detecting} use a similar idea of rigidity constraint and propose a rigidity geometry testing for moving rigid object matching.
\yuheng{Previous research utilized rigidity assumption to segment moving rigid objects, while in this paper, we use rigidity constraint to recover objects' structure.} 

To model rigid object, SLAM++ \cite{salas2013slam++} introduced pre-defined CAD models into the object matching and pose optimization.
QuadricSLAM \cite{nicholson2018quadricslam} utilize dual-quadrics as 3D object representation, to represent the orientation and scale of object landmarks.
Yang and Scherer \cite{yang2019cubeslam} propose a monocular object SLAM system named CubeSLAM for 3D cuboid object detection and multi-view object SLAM.
As mentioned earlier, the above methods can only model simple rigid objects, e.g., cubes, while we target more general objects, i.e., \emph{articulated objects}, which can cover common dynamic objects such as vehicles and humans.

\section{Methodology}

\subsection{Background and Notation}
Visual SLAM in static environments is often formulated as a factor graph optimization \cite{kaess2008isam}.
The objective \eqref{eq:static obs} is to find the robot state $x_k\in X$, $k \in [0, n_x]$ and the static landmarks ${p_i}\in P_s$, $i \in [0, n_{p_s}]$ that best fit the observation of the landmarks $z_{k}^i\in Z$, where $n_x$ denotes the total number of the robots' state and $n_{p_s}$ denotes the number of the static landmarks. This is often based on a reprojection error minimization \: $e_{i,k}=\|h(x_k,p^i)-z_{k}^i\|$ with:
\begin{equation}\label{eq:static obs}
X^*, P^* = \argmin_{\{X,P_s\}} \sum_{i,k} e_{i,k}^T \Omega_{i,k}^{-1}  e_{i,k}
\end{equation}
where $h(x_k,p_i)$ denotes the 3D points observation function and $\Omega_{i,k}$ denotes the observation covariance matrix.

In dynamic SLAM, the reprojection error $e_p$ of dynamic feature points is also considered:
\begin{equation}\label{dyobservation}
e_p = \|h({x_k}, {_lp^i_k}) - {}_lz^i_k\|,
\end{equation}
where ${_lp_k^i}\in P_d$ are the dynamic points and $_lz_{k}^i$ are the corresponding observation of dynamic points.

\subsection{Rigidity Constraint}\label{Sec:Rigidity}

Let $s^{ij}$ be the segment length between two feature points $_lp_k^i$ and $_lp_k^j$, the rigidity constraint is that $s^{ij}$ is invariant over time, i.e, $s^{ij}_k=s^{ij}_{k+1}$, if $_lp_k^i$ and $_lp_k^j$ are on the same rigid part of an articulated object, as shown in \fref{Fig:point-segment model}.
Inspired by this, we model the dynamic articulated object using a rigidity constraint, and thus we can define the rigidity error $e_r$ as

\begin{equation}\label{eq:rigidity}
e_r = \|\|_lp_k^i - \prescript{}{l}{p}_k^j\| - s^{ij}\|.
\end{equation}

\begin{figure}[t]
	\centering
	    \subfigure[Rigidity Constraint Factor Graph ]{
    	\includegraphics[width=0.6\linewidth]{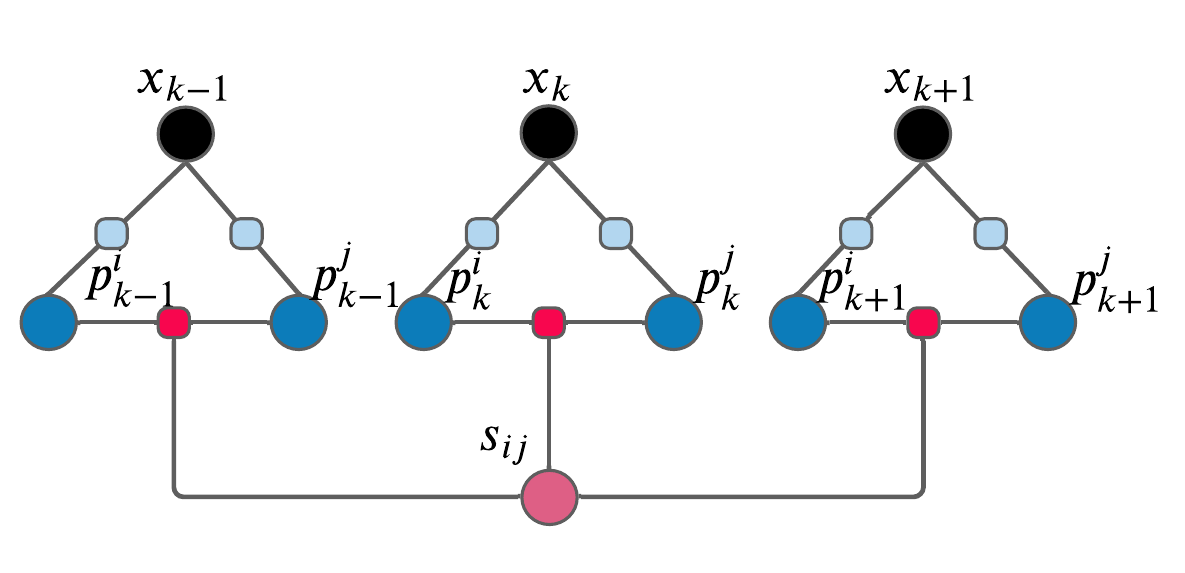}
    	\label{Fig:factorgraph_rigidity}
	}
	\subfigure[Human Rigidity]{
    \includegraphics[width=0.3\linewidth]{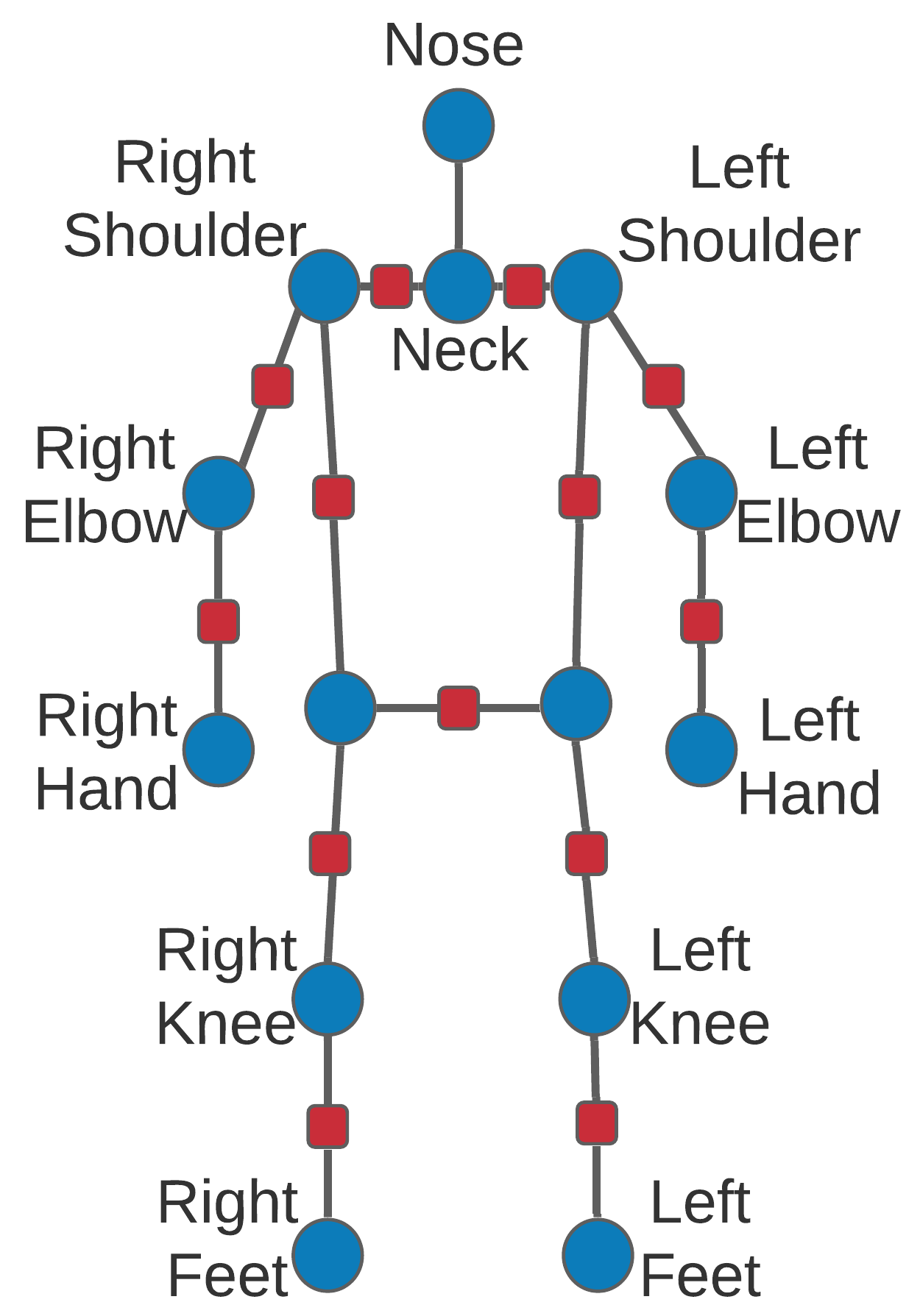}
	\label{Fig:human model}
    }
    \\
        \subfigure[Motion Constraint Factor Graph]{
	\includegraphics[width=0.6\linewidth]{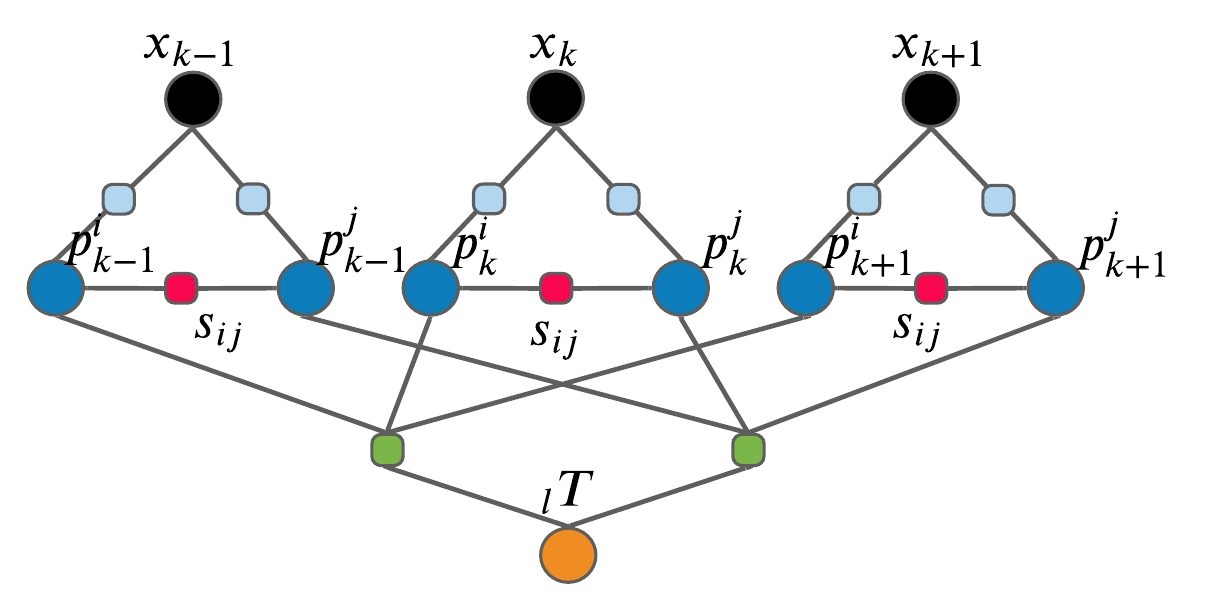}
	\label{Fig:factorgraph_motion}
	}

	\caption{(a) Factor graph of the rigidity constraint. Black nodes represent the camera pose, blue nodes the dynamic points, and red nodes indicate the rigid segment length. Cyan and red rectangles represent the measurements of points and rigidity consequently. (c) Human can be modeled with point and segment based on the body parts' rigidity. (b) Factor graph of the motion constraint. The orange node is the estimated motion and the green rectangles denote the motion constraints }
\end{figure}

\fref{Fig:factorgraph_rigidity} shows the factor graph of the rigidity constraint, where the length of segment $s_{ij}$ is invariant after the motion.
The benefits to involving the rigidity error \eqref{eq:rigidity} are two-fold.
First, it offers a temporal geometric constraint for dynamic points, which is able to correct the scale and 3D structure of dynamic objects. Second, it provides a geometric check, which eliminates the incorrectly matched points.

We model humans as a special articulated object shown in \fref{Fig:human model}, where each human can be described by 14 key points, including nose, shoulders, elbows, hands, waists, knee, feet, etc. In the experiments, we detect the human key points using the off-the-shelf algorithm Alpha-Pose \cite{fang2017rmpe}.

\begin{figure*}[t]
	\centering
    \includegraphics[width=0.95\linewidth]{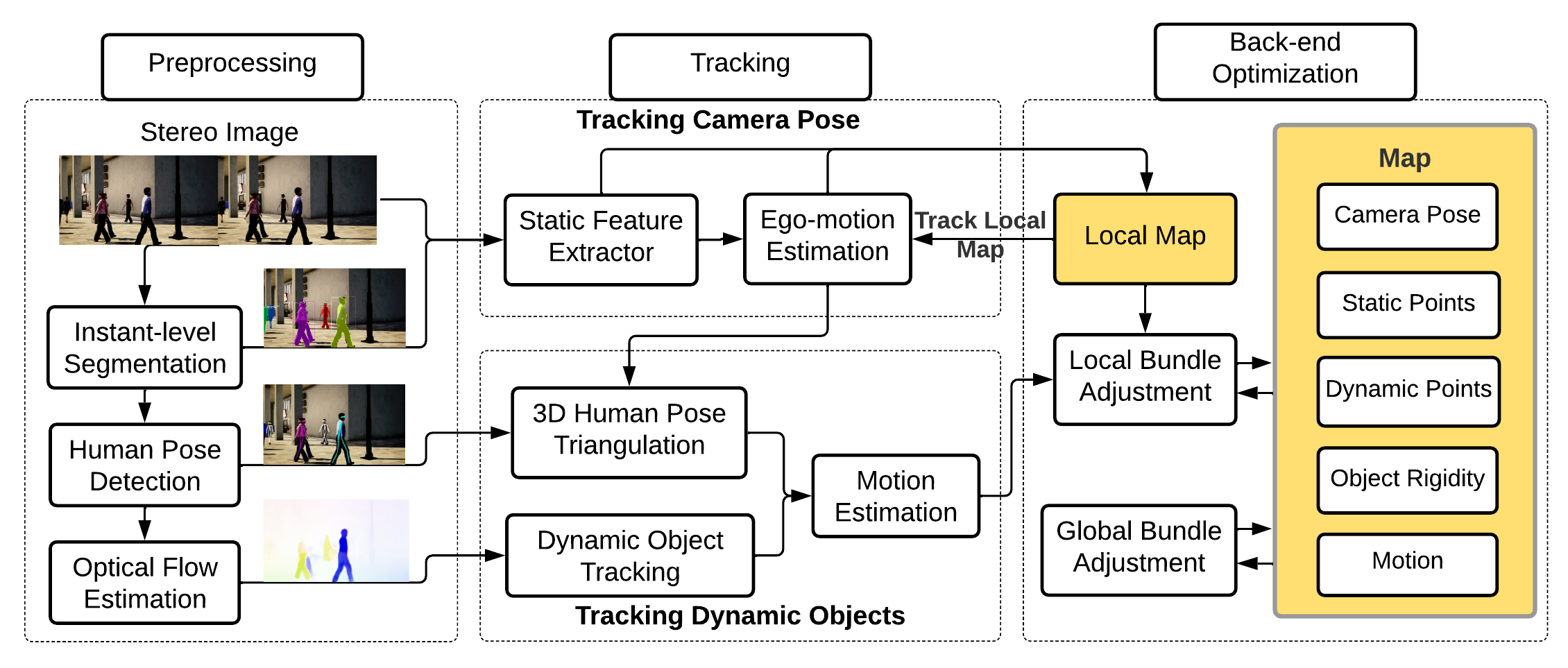}
    \vspace{-10pt}
			\caption{The framework of AirDOS, which is composed of three modules, i.e., pre-processing, tracking, and back-end optimization.}
	\vspace{-6pt}
	\label{Fig:system}
\end{figure*}

\subsection{Motion Constraint}\label{Sec:Motion}
We adopt the motion constraint from \cite{henein2018exploiting} which does not need a prior geometric model. For every feature point on the same rigid part of an articulated object $l$, we have
\begin{equation}
_l \bar{p}_{k+1}^i = \prescript{}{l}{T} \prescript{}{l}{\bar{p}}_{k}^i,
\end{equation}
where $_lT \in SE(3)$ is a motion transform associated with the object $l$ and $\bar{\cdot}$ indicates homogeneous coordinates.
% , with $_lT \in T$, the set of motion. 
Therefore, we can define the loss function for motion constraint as:
\begin{equation}\label{eq:motion}
e_m = || \prescript{}{l}{\bar{p}}_{k+1}^i - \prescript{}{l}{T} \prescript{}{l}{\bar{p}}_{k}^i||.
\end{equation}

The motion constraint simultaneously estimates the objects' motion $_lT$ and enforces each point $_lp_k^i$ to follow the same motion pattern \cite{henein2018exploiting}. \yuheng{This motion model $_lT$ assumes that the object is rigid, thus, for articulated objects, we apply the motion constraint on each rigid part of articulated object.} In \fref{Fig:factorgraph_motion} we show the factor graph of the motion constraint.

In highly dynamic environments, even if we filter out the moving objects, the tracking of static features is easily interrupted by the moving objects.
By enforcing the motion constraints, dynamic objects will be able to contribute to the motion estimation of the
camera pose.
Therefore, when the static features are not reliable enough, moving objects can correct the camera pose estimation, preventing tracking loss.

\subsection{Bundle Adjustment}\label{sec:BA}
The bundle adjustment (BA) jointly optimizes the static points $p^i$, dynamic points $_lp^i_k$, segments $s^{ij}$, camera poses $x_k$ and dynamic object motions $_lT$. This can be formulated as the factor graph optimization:
\begin{multline}
X^*, P^*, S^*, T^* = \argmin_{\{X,P,S,T\}} e_r^T\Omega_{i,j}^{-1}e_r + \\ e_m^T\Omega_{i,l}^{-1}e_m +
e_p^T\Omega_{i,k}^{-1}e_p,\label{eq: Full BA}
\end{multline}
where $P$ is the union set of $P_s$ and $P_d$. This problem can be solved using the Levenberg-Marquardt algorithms.

\section{System Overview}\label{sec: system}
We propose the framework AirDOS in \fref{Fig:system} for dynamic stereo visual SLAM, which consists of three modules, pre-processing, tracking, and back-end bundle adjustment.

In pre-processing and tracking modules, we first extract ORB features \cite{rublee2011orb} and perform an instance-level segmentation \cite{maskrcnn} to identify potential moving objects.
We then estimate the initial ego-motion by tracking the static features.
For articulated objects like humans, we perform Alpha-Pose \cite{fang2017rmpe} to extract the human key points and calculate their 3D positions by triangulating the corresponding key points from stereo images.
We then track the moving humans using the optical flow generated by PWC-net \cite{sun2018pwc}.
The tracking module provides a reliable initialization for the camera pose and also the object poses of dynamic objects.

In the back-end optimization, we construct a global map consisting of camera poses, static points, dynamic points, and the motion of objects.
% We also perform bundle adjustment with RANSAC \cite{fischler1981random} using rigidity and motion constraints on the local co-visibility graph, built from the shared observation of moving objects.
\yuheng{We perform local bundle adjustment with dynamic objects in the co-visibility graph \cite{mei2010closing} built from the co-visible landmarks for the sake of efficiency. Similar to the strategy of RANSAC, we eliminate the factors and edges which contribute a large error based on the rigidity constraint \eqref{eq:rigidity} and motion constraint \eqref{eq:motion}.}
This process helps to identify the mismatched or falsely estimated human poses.
Visual SLAM algorithms usually only perform bundle adjustment on selected key-frames due to the repeated static feature observations. \yuheng{However, in highly dynamic environments, like the ones presented in this paper, this might easily result in loss of dynamic object tracking}, therefore we perform bundle adjustment on every frame to capture the full trajectory.

\section{Experiments}

\begin{table}[t]
\caption{Performance on KITTI datasets based on ATE (\meter).}
\vspace{-2pt}
\centering
\small
% \begin{threeparttable}
\resizebox{0.45\textwidth}{!}{
\begin{tabular}{ccccc}
\toprule
\multirow{2}{*}{\textbf{Sequence}} & \multicolumn{2}{c}{\textbf{W/ Mask}} & \multicolumn{2}{c}{\textbf{W/O Mask}}\\
  & ORB-SLAM & AirDOS & ORB-SLAM & AirDOS \\ 
\midrule

 Test 18 & 0.933 & 0.934 & 0.937& 0.948 \\ 
 Test 28 & 2.033 & 2.027 & 2.031& 2.021 \\ 
 Train 13 & 1.547 & 1.618 & 1.551& 1.636 \\ 
 Train 14 & 0.176 & 0.172 & 0.174& 0.169 \\ 
 Train 15 & 0.240 & 0.234 & 0.240& 0.234 \\ 
 Train 19 & 2.633 & 2.760 & 2.642& 2.760 \\ 
\bottomrule
\end{tabular}
}
\label{tab: KITTI}
\vspace{-8pt}
\end{table}

\subsection{Metric, Baseline, and Implementation}
We use the Absolute Translation Error (ATE) to evaluate our algorithm. Our method is compared against the state-of-the-art methods, ORB-SLAM \cite{mur2015orb} (1) with and (2) without the masking of potential dynamic objects, and RGB-D dynamic SLAM algorithm \cite{zhang2020vdo}.
\yuheng{Similar to the setup described in \sref{sec: system}, we modified the ORB-SLAM to perform BA on every frame with the observation from dynamic features, so as to capture the full trajectory of the moving objects.  In the experiment, we applied the same parameters to AirDOS and ORB-SLAM, i.e., the number of feature points extracted per frame, the threshold for RANSAC, and the covariance of the reprojection error.}
% In the experiment setup, we applied the same parameters to AirDOS and ORB-SLAM. 
% Our algorithm is implemented based on ORB-SLAM \cite{mur2015orb}.
% , which includes a front-end stereo visual odometry and a back-end bundle adjustment based on static ORB features.
%

\begin{figure}[t]
	\centering
    \includegraphics[width=\linewidth]{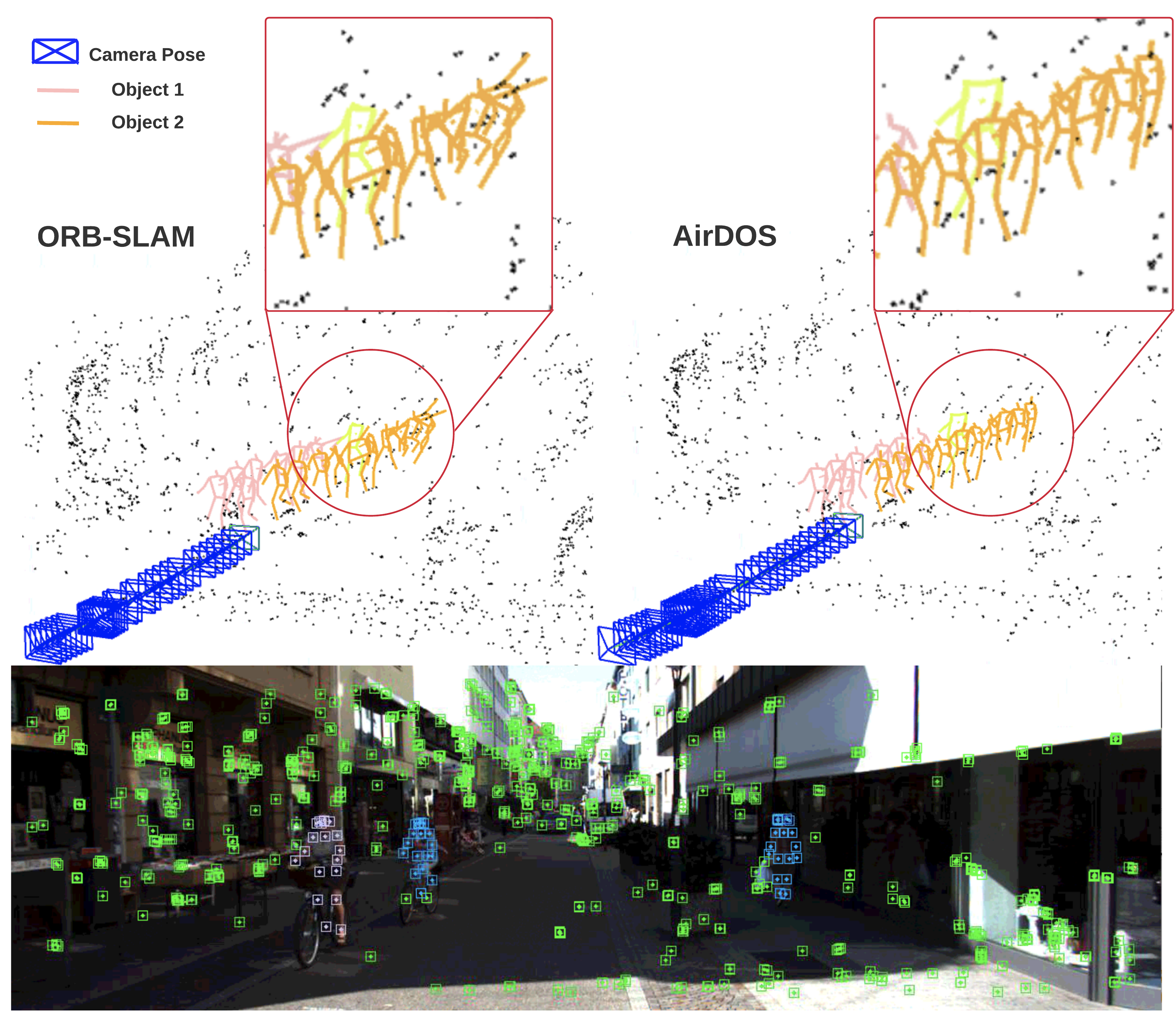}
    % \label{Fig:kitti}
	\caption{Qualitative analysis of the KITTI Tracking datasets in training 19. Applying rigidity constraint and motion constraint improve the estimation of the objects' structure.
	}
	\label{Fig: KITTIQua}
	\vspace{-8pt}
\end{figure}

\begin{table*}[t]
    \centering
    \small
    % \resizebox{0.8\textwidth}{!}{
    \begin{threeparttable}
        \caption{Experiments on Tartan-Air dataset with and without Mask}
        \begin{tabular}{C{0.08\linewidth}C{0.08\linewidth}|C{0.12\linewidth}C{0.12\linewidth}C{0.14\linewidth}|C{0.12\linewidth}C{0.12\linewidth}}
            \toprule
            \multirow{2}{*}{\textbf{Datasets}} & \multirow{2}{*}{\textbf{Sequence}} & \multicolumn{3}{c|}{\textbf{W/ Mask}} & \multicolumn{2}{c}{\textbf{W/O Mask}}\\
            
            && AirDOS & ORB-SLAM  &VDO-SLAM\cite{zhang2020vdo} & AirDOS & ORB-SLAM\\ 
            \midrule
            \multirow{2}{*}{\parbox{2cm} {\centering Standing Human}} & \rom{1} & \textbf{0.0606} & 0.0788 & 0.0994 & \textbf{0.0469}  &0.1186 \\ 
             & \rom{2} & 0.0193 & \textbf{0.0060} & 0.6129 & - & -\\
             \midrule
            \multirow{3}{*}{\parbox{2cm} {\centering Road Crossing (Easy)}}  & \rom{3} & 0.0951   & \textbf{0.0657} & 0.3813 & \textbf{0.0278}  & 0.0782 \\ 
             & \rom{4} & 0.0331 & \textbf{0.0196} & 0.3879 & 0.1106 & \textbf{0.0927} \\  
             & \rom{5} & 0.0206 & \textbf{0.0148} & 0.2175 & \textbf{0.0149} & 0.0162\\ 
             \midrule
             \multirow{2}{*}{\parbox{2cm} {\centering Road Crossing (Hard)}} & \rom{6} & \textbf{0.2230} & 1.0984 & 0.2400 & \textbf{3.6700} & 4.3907 \\ 
            %  & 6 & 1.0013 & 1.1237 &  \textbf{0.2485}  & - & -  \\ 
             & \rom{7} & \textbf{0.5625} & 0.8476 & 0.6628 & \textbf{1.1572} & 1.4632\\ 

             \midrule
            %  \multicolumn{2}{c|}{\textbf{Overall}} & \textbf{0.2483} & 0.4105 & 0.3563 & \textbf{0.8379} &1.0226\\
          \multicolumn{2}{c|}{\textbf{Overall}} & \textbf{0.1449} & 0.3044 & 0.3717 & \textbf{0.8379} &1.0226\\
            \bottomrule
            \end{tabular}

            \begin{tablenotes}[normal,flushleft]
            \item {Results show Absolute Trajectory Error (ATE) in meter (m). `-' means that SLAM failed in this sequence.}
            %\item {}
            \end{tablenotes}
        \label{tab:tartanair}
    \end{threeparttable}
                % }
                \vspace{-8pt}
\end{table*}

\subsection{Performance on KITTI Tracking Dataset}
The KITTI Tracking dataset \cite{geiger2013vision} contains 50 sequences (29 for testing, 21 for training) with multiple moving objects. 
We select 6 sequences that contain moving pedestrians. For evaluation, we generate the ground truth using IMU and GPS.
As shown in \tref{tab: KITTI}, the ATEs of both our method and ORB-SLAM are small in all sequences, which means that both methods perform well in these sequences.
The main reason is that the moving objects are relatively far and small, and there are plentiful static features in these sequences. Moreover, most sequences have a simple translational movement, which makes these cases very simple.

Although the camera trajectory is similar, our algorithm recovers a better human model as shown in \fref{Fig: KITTIQua}. The ORB-SLAM generates noisy human poses when the human is far away from the camera. That's because the rigidity constraint helps to recover the structure of the moving articulated objects. Also, the motion constraint can improve the accuracy of the dynamic objects' trajectory.
Given the observation from the entire trajectory, our algorithm recovers the human pose and eliminates the mismatched dynamic feature points.

\begin{figure}[t]
	\centering
    \subfigure[Standing Human]{
    	\includegraphics[width=0.46\linewidth]{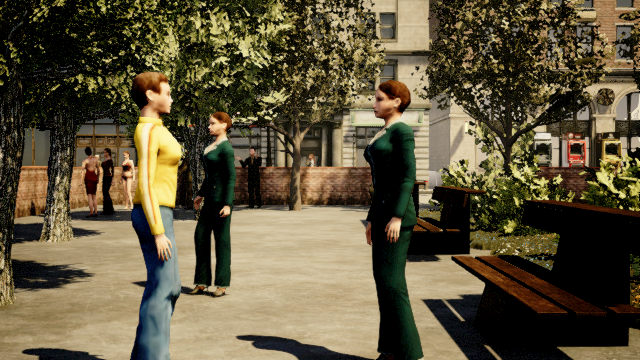}
    	\label{Fig:standing}
	}
	\subfigure[Road Crossing]{
    	\includegraphics[width=0.46\linewidth]{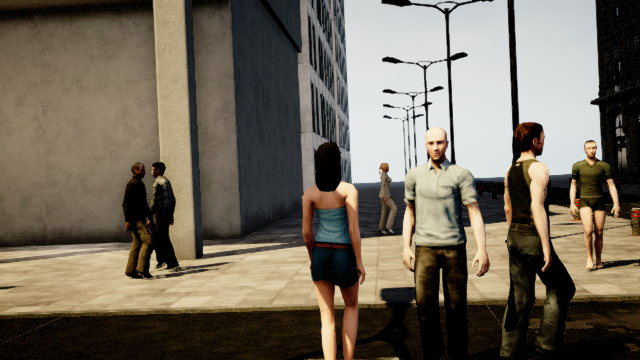}
    	\label{Fig:roadcrossing}
	}

	\caption{(a) Example of the Tartan-Air datasets, where almost every one is standing. (b) Example of moving humans in road crossing.}
	\label{fig:tartan-air-shibuya}
		\vspace{-8pt}
\end{figure}

\subsection{Performance on TartanAir Shibuya Dataset}

We notice that the moving objects in KITTI dataset only cover a small field of view.
To address the challenges of the highly dynamic environment, we build the TartanAir Shibuya dataset as shown in \fref{fig:tartan-air-shibuya}, and demonstrate that our method outperforms the existing dynamic SLAM algorithms in this benchmark. 
Our previous work TartanAir \cite{wang2020tartanair} is a very challenging visual SLAM dataset consisting of binocular RGB-D sequences together with additional per-frame information such as camera poses, optical flow, and semantic annotations.
In this paper, we use the same pipeline to generate TartanAir Shibuya, which is to simulate the world's most busy road intersection at Shibuya Tokyo shown in \fref{fig:motivation}.
It covers much more challenging viewpoints and diverse motion patterns for articulated objects than TartanAir.

We separate the TartanAir Shibuya dataset into two groups: Standing Humans in \fref{Fig:standing} and Road Crossing in \fref{Fig:roadcrossing} with easy and difficult categories. Each sequence contains 100 frames and more than 30 tracked moving humans. In the sequences of Standing Human, most of the humans standstill, while few of them move around the space. In Road Crossing, there are multiple moving humans coming from different directions. For the difficult sequences, dynamic objects often enter the scene abruptly, in which the visual odometry of traditional methods will fail easily.

\begin{figure}[t]
	\centering
    	\includegraphics[width=\linewidth]{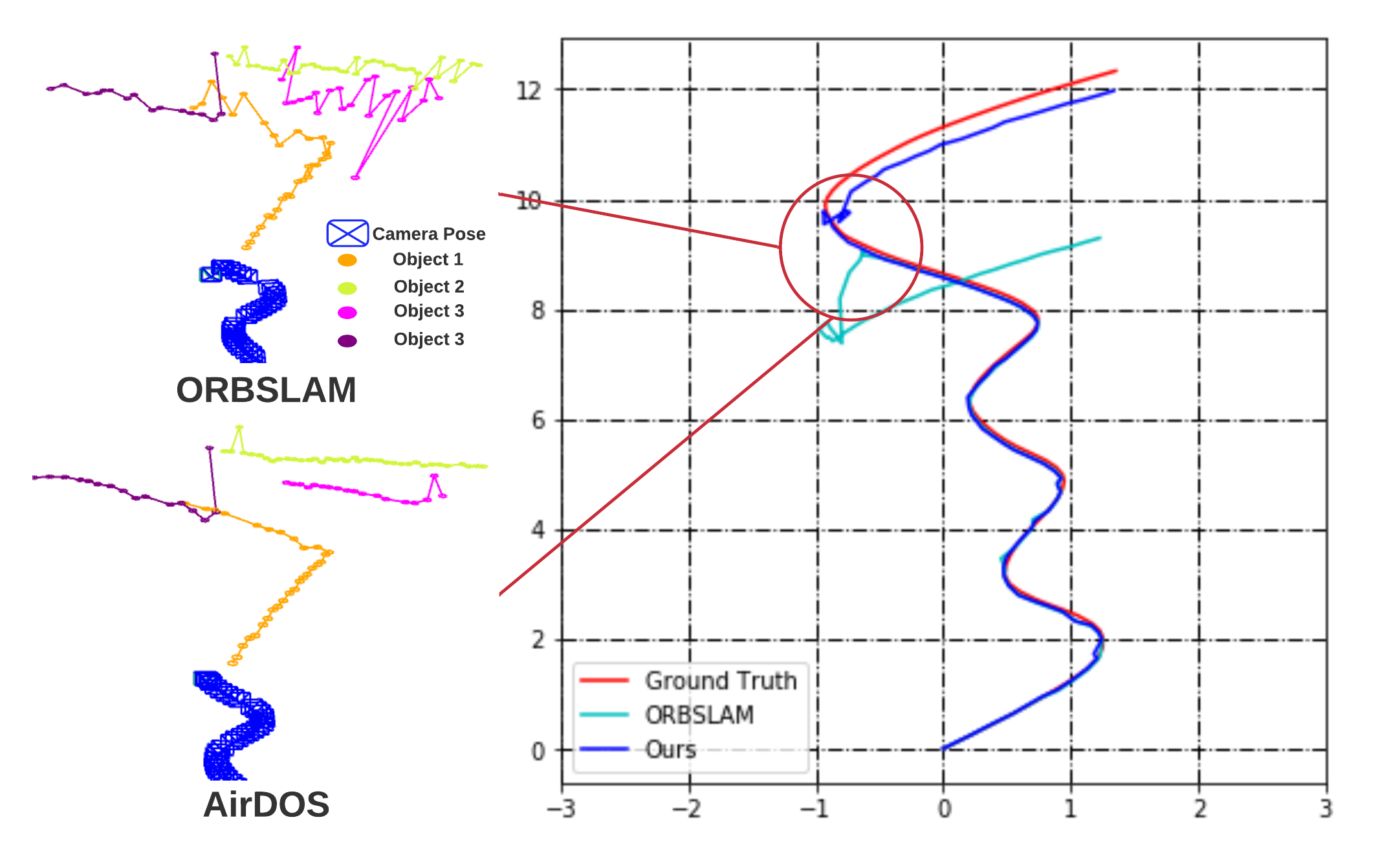}
	\caption{Qualitative analysis of the TartanAir sequence \rom{4}. The moving objects tracked by the ORB-SLAM is noisy, while our proposed method generate a smooth trajectory. We present that dynamic objects and the camera pose can benefits each other in visual SLAM.}
	\label{fig: tartanquat}
	\vspace{-8pt}
\end{figure}

\subsubsection{Evaluation}

\begin{table*}[t]
    \caption{Ablation study on simulated dataset.}
    \label{tab:ablation study}
    \centering
    \resizebox{\textwidth}{!}{
    \begin{tabular}{ccccccccccccc|ccc}
        \toprule
        \multirow{2}{*}{\textbf{Groups}} & \multicolumn{3}{c}{\textbf{\rom{1}}} & \multicolumn{3}{c}{\textbf{\rom{2}}} & \multicolumn{3}{c}{\textbf{\rom{3}}} & \multicolumn{3}{c|}{\textbf{\rom{4}}} & \multicolumn{3}{c}{\textbf{Overall}} \\
        & RPE-R & RPE-T & ATE & RPE-R & RPE-T & ATE & RPE-R & RPE-T & ATE & RPE-R & RPE-T & ATE & RPE-R & RPE-T & ATE\\
        \midrule
        Before BA          & 0.4898 & 15.991 & 83.441 & 0.6343 & 17.728 & 109.968 & 1.1003 & 21.070 & 138.373 & 0.7925 & 19.242 & 168.312 & 0.7908 & 18.8328 & 125.024  \\
        BA w/ static point & 0.0989 & 3.3184 & 15.002 & 0.1348 & 3.7146 & 25.796 & 0.2028 & 4.2522 & 17.085 & 0.1389 & 3.5074 & 35.521 & 0.1537 & 3.7540 & 23.351   \\
        BA w/o  motion     & 0.0988 & 3.3176 & 15.019 & 0.1349 & 3.7176 & 25.708 & 0.2035 & 4.2631 & 16.985 & 0.1388 & 3.5069 & 35.269 & 0.1538 & 3.7565 & 23.245  \\
        BA w/o  rigidity   & 0.0962 & 3.2245 & 14.881 & 0.1282 & 3.4984 & 25.704 & 0.1871 & 4.0387 & 16.921 & 0.1226 & 3.2397 & 35.426 & 0.1410 & 3.5148 & 23.233 \\
        BA in Equation \eqref{eq: Full BA} %AirDOS
        & \textbf{0.0958}     & \textbf{3.2177} & \textbf{14.879} & \textbf{0.1276} & \textbf{3.4824} & \textbf{25.703} & \textbf{0.1870} & \textbf{4.0372} & \textbf{16.914} & \textbf{0.1215} & \textbf{3.2227}  & \textbf{35.412} & \textbf{0.1407} & \textbf{3.5085} & \textbf{23.227}\\
        \bottomrule
    \end{tabular}
    }
    \begin{tablenotes}[normal]
    \item {Results show RPE-T and ATE in centimeter (\centi\meter) and RPE-R in degree (\degree).}
    \end{tablenotes}
    \vspace{-8pt}
\end{table*}

To test the robustness of our system when the visual odometry is interrupted by dynamic objects or in cases where the segmentation might fail due to indirect occlusions such as illumination changes, we evaluate the performance in two settings: with and without masking the dynamic features during ego-motion estimation.

As shown in the \tref{tab:tartanair}, with human masks, our algorithm obtains a 39.5\% and 15.2\% improvements compared to ORB-SLAM \cite{mur2015orb} and VDO-SLAM \cite{zhang2020vdo} in the overall performance.
In Sequence \rom{2}, \rom{4} and \rom{5}, both ORB-SLAM and our algorithm show a good performance, where all ATEs are lower than 0.04. We notice that the performance of VDO-SLAM is not as good as ORB-SLAM. 
This may be because that VDO-SLAM relies heavily on the optical flow for feature matching, it is likely to confuse background features with dynamic features.

Our algorithm also outperforms ORB-SLAM without masking the potential moving objects. 
As shown in the sequence \rom{1}, \rom{3}, \rom{5}, and \rom{6} of \tref{tab:tartanair}, our method obtains a higher accuracy than ORB-SLAM by 0.0717, 0.050, 0.721 and 0.306. Overall, we achieve an improvement of 18.1\%.
That's because moving objects can easily lead the traditional visual odometry to fail, but we take the observations from moving articulated objects to rectify the camera poses, and filter out the mismatched dynamic features. 

It can be seen that in \fref{fig: tartanquat}, ORB-SLAM was interrupted by the moving humans and failed when making a large rotation. By tracking moving humans, our method outperforms ORB-SLAM when making a turn. Also, a better camera pose estimation can in turn benefit the moving objects' trajectory. As can be seen, the objects' trajectories generated by ORB-SLAM are noisy and inconsistent, while ours are smoother.
In general, the proposed motion constraint and rigidity constraint have a significant impact on the difficult sequences, where ORB-SLAM outputs inaccurate trajectories due to dynamic objects. 

\section{Ablation Study}
We perform an ablation study to show the effects of the introduced rigidity and motion constraints.
% Specifically, we demonstrate that the two constraints make the camera pose estimation and dynamic object estimation mutually beneficial via bundle adjustment (BA).
Specifically, we demonstrate that the motion constraint and rigidity constraint inprove the camera pose estimation via bundle adjustment.

\subsection{Implementation}\label{sec: ablation_implement}
We simulate dynamic articulated objects that follow a simple constant motion pattern, and initialize the robot's state with Gaussian noise of $\sigma = 0.05\meter$ on translation $\sigma = 2.9\degree$ on rotation.
We also generate static features around the path of the robot, and simulate a sensor with a finite field of view.
The measurement of point also has a noise of $\sigma = 0.05\meter$.
We generate 4 groups of sequences with different lengths and each group consists of 10 sequences that are initialized with the same number of static and dynamic features. We set the ratio of static to dynamic landmarks as 1:1.8.

\subsection{Results}
We evaluate the performance of (a) bundle adjustment with static features only, (b) bundle adjustment without motion constraint, (c) bundle adjustment without rigidity constraint, and (d) bundle adjustment with both the motion constraint and rigidity constraint.
We use the Absolute Translation Error (ATE) and Relative Pose Error of Rotation (RPE-R) and Translation (RPE-T) as our evaluation metrics.

As shown in \tref{tab:ablation study}, both motion and rigidity constraints are able to improve the camera pose estimation, while the best performance is obtained when the two constraints are applied together.
An interesting phenomenon is that rigidity constraint can also benefit the objects' trajectory estimation. 
In Groups \rom{1}, we evaluate the estimation of dynamic points with setting (b), (c), and (d), with 100 repeated experiments.
We find that the ATE of dynamic object feature points in setting (c) is 5.68 $\pm$ 0.30 lower than setting (b), while setting (d) is 5.71 $\pm$ 0.31 lower than (b).
This is because the motion constraint assumes that every dynamic feature on the same object follows the same motion pattern, which requires the object to be rigid.
From another point of view, the rigidity constraint provides a good initialization to the objects' 3D structure, and so indirectly improves the estimation of the objects' trajectory. 
In general, the ablation study proves that applying motion and rigidity constraints to dynamic articulated objects can benefit the camera pose estimation.

\subsection{Computational Analysis}
\yuheng{
Finally, we evaluate the running time of the rigidity constraint and motion constraint in the optimization. The back-end optimization is implemented in C++ with a modified g2o \cite{kummerle2011g} solver. With the same setup as section \ref{sec: ablation_implement}, we randomly initialized 10 different sequences with 18 frames. In each frame, we can observe 8 static landmarks, and 12 dynamic landmarks from one moving object. In \tref{tab:time analysis1}, We show the (i) convergence time (ii) runtime per iteration of group \rom{1} in the ablation study. Our method takes 53.54 (mSec) to converge, which is comparable to 39.22 (mSec) from the optimization with re-projection error only.
}

\yuheng{In this paper, semantic mask \cite{maskrcnn} and human poses \cite{fang2017rmpe} are pre-processed as an input to the system. The experiment are carried out on an Intel Core i7 with 16GB RAM.}
\vspace{-2pt}
\begin{table}[h]
    \caption{Time Analysis of Bundle Adjustment }
    \label{tab:time analysis1}
    \centering
    \small
    \resizebox{0.48\textwidth}{!}{
    \begin{tabular}{ccc}
        \toprule
        % \multirow{2}{*}{\textbf{Groups}} & \multicolumn{2}{c}{\textbf{\rom{1}}} & \multicolumn{2}{c}{\textbf{\rom{2}}} & \multicolumn{2}{c}{\textbf{\rom{3}}} & \multicolumn{2}{c}{\textbf{\rom{4}}}  \\
         & Convergence Time (mSec) & Runtime/iter (mSec)   \\
         \midrule
        
        BA w/ reprojection error & 39.22& 4.024  \\
        BA w/o Rigidity &45.47&4.078 \\
        BA w/o Motion&45.37&4.637\\
        BA in Equation \eqref{eq: Full BA}&53.54&4.792\\

        \bottomrule
        % \multicolumn{9}{c}{{Groups}} \\
    \end{tabular}
    }
    
    % \begin{tablenotes}[normal]
    % \item {Results show time for msec}
    % \end{tablenotes}
\end{table}
\vspace{-8pt}

\section*{Conclusion}

In this paper, we introduce the rigidity constraint and motion constraint to model dynamic articulated objects. We propose a new pipeline, AirDOS for stereo SLAM which jointly optimizes the trajectory of dynamic objects, map of the environment, and camera poses, improving the robustness and accuracy in dynamic environments. We evaluate our algorithm in KITTI tracking and TartanAir Shibuya dataset, and demonstrate that camera pose estimation and dynamic objects can benefit each other, \yuheng{especially when there is an aggressive rotation or static features are not enough to support the visual odometry.}

\bibliographystyle{./bibliography/IEEEtran}
\bibliography{./bibliography/papers}

\end{document}